\pdfoutput=1

\documentclass[11pt]{article}

\usepackage[]{acl}

\usepackage{times}
\usepackage{latexsym}

\usepackage[T1]{fontenc}

\usepackage[utf8]{inputenc}

\usepackage{microtype}

\usepackage{hyperref}
\usepackage{url}
\usepackage{xspace}
\usepackage{amsmath}
\usepackage{amsthm}
\usepackage{booktabs}
\usepackage{multirow}
\usepackage{multicol}
\usepackage{ifthen}
\usepackage{graphicx}
\usepackage{multirow}
\usepackage{soul,xcolor,colortbl}
\usepackage{color,soul}
\usepackage[T1]{fontenc}
\usepackage{pifont}%
\usepackage[scaled=.85]{beramono}
\usepackage{colortbl}
\usepackage{subcaption}

\usepackage{amsmath,amsfonts,bm}

\def\eqref#1{equation~\ref{#1}}

\def\1{\bm{1}}

\DeclareMathAlphabet{\mathsfit}{\encodingdefault}{\sfdefault}{m}{sl}
\SetMathAlphabet{\mathsfit}{bold}{\encodingdefault}{\sfdefault}{bx}{n}

\DeclareMathOperator*{\argmax}{arg\,max}

\newcommand\hh[1]{\textcolor{blue}{[HH: #1]}}
\newcommand\nj[1]{\textcolor{brown}{[NJ: #1]}}

\renewcommand\hh[1]{}
\renewcommand\nj[1]{}

\definecolor{cadd}{rgb}{0, 0.57, 0.34}
\definecolor{cdel}{rgb}{0.9, 0.22, 0.16}
\definecolor{cstrik}{rgb}{0.9, 0.6, 0.6}
\definecolor{lightapricot}{rgb}{0.99, 0.84, 0.69}

\newcolumntype{L}[1]{>{\raggedright\let\newline\\\arraybackslash\hspace{0pt}}m{#1}}
\newcolumntype{C}[1]{>{\centering\let\newline\\\arraybackslash\hspace{0pt}}m{#1}}
\newcolumntype{R}[1]{>{\raggedleft\let\newline\\\arraybackslash\hspace{0pt}}m{#1}}

\setstcolor{cstrik}

\newcommand\sD{\ensuremath{\mathcal{D}}}

\newcommand\sF{\ensuremath{\mathcal{F}}}

\newcommand\sN{\ensuremath{\mathcal{N}}}

\newcommand\BE{\ensuremath{\mathbb{E}}}

\newcommand\BR{\ensuremath{\mathbb{R}}}

\newcommand\pb[1]{\ensuremath{\left[ #1 \right]}} 

\newcommand\refeqn[1]{(\ref{eqn:#1})}

\newcommand\refsec[1]{Section~\ref{sec:#1}}

\newcommand\reffig[1]{Figure~\ref{fig:#1}}

\ifthenelse{\isundefined{\definition}}{\newtheorem{definition}{Definition}}{}
\ifthenelse{\isundefined{\assumption}}{\newtheorem{assumption}{Assumption}}{}
\ifthenelse{\isundefined{\hypothesis}}{\newtheorem{hypothesis}{Hypothesis}}{}
\ifthenelse{\isundefined{\proposition}}{\newtheorem{proposition}{Proposition}}{}
\ifthenelse{\isundefined{\theorem}}{\newtheorem{theorem}{Theorem}}{}
\ifthenelse{\isundefined{\lemma}}{\newtheorem{lemma}{Lemma}}{}
\ifthenelse{\isundefined{\corollary}}{\newtheorem{corollary}{Corollary}}{}
\ifthenelse{\isundefined{\alg}}{\newtheorem{alg}{Algorithm}}{}
\ifthenelse{\isundefined{\example}}{\newtheorem{example}{Example}}{}

\newcommand{\ctrltag}[1]{\texttt{#1}\xspace}
\newtheorem{prop}{Proposition}

\title{An Investigation of the (In)effectiveness of Counterfactually Augmented Data}

\author{
 Nitish Joshi$^1$ \ \ \ \ \ \ \ \ \ \ \ \ \  He He$^{1,2}$  \\
 $^1$Department of Computer Science, New York University\\
 $^2$Center for Data Science, New York University \\
  {\texttt{nitish@nyu.edu}, \texttt{hhe@nyu.edu}} \\
}

\begin{document}
\maketitle
\begin{abstract}
    While pretrained language models achieve excellent performance on natural language understanding benchmarks, 
they tend to rely on spurious correlations and generalize poorly to out-of-distribution (OOD) data.
Recent work has explored using counterfactually-augmented data (CAD)---data generated by minimally perturbing examples to flip the ground-truth label---to identify robust features that are invariant under distribution shift.
However, empirical results using CAD during training for OOD generalization have been mixed.
To explain this discrepancy, through a toy theoretical example and empirical analysis on two crowdsourced CAD datasets, we show that: (a) while features perturbed in CAD are indeed robust features, it may prevent the model from learning \emph{unperturbed} robust features; and (b) CAD may exacerbate existing spurious correlations in the data. Our results thus show that the lack of perturbation diversity limits CAD's effectiveness on OOD generalization,
calling for innovative crowdsourcing procedures to elicit diverse perturbation of examples.
\end{abstract}

\section{Introduction}
\label{sec:intro}

Large-scale datasets have enabled tremendous progress in natural language understanding (NLU) ~\citep{rajpurkar-etal-2016-squad, wang2018glue} with the rise of pretrained language models~\citep{devlin-etal-2019-bert, peters-etal-2018-deep}.
Despite this progress, there have been numerous works showing that models rely on spurious correlations in the datasets,
i.e. heuristics that are effective on a specific dataset but do not hold in general~\citep{mccoy-etal-2019-right, naik-etal-2018-stress, Wang2020IdentifyingSC}.
For example, BERT~\citep{devlin-etal-2019-bert} trained on MNLI~\citep{williams2017broad} learns the spurious correlation between world overlap and entailment label.

\begin{figure*}
    \centering
    \includegraphics[scale=0.6]{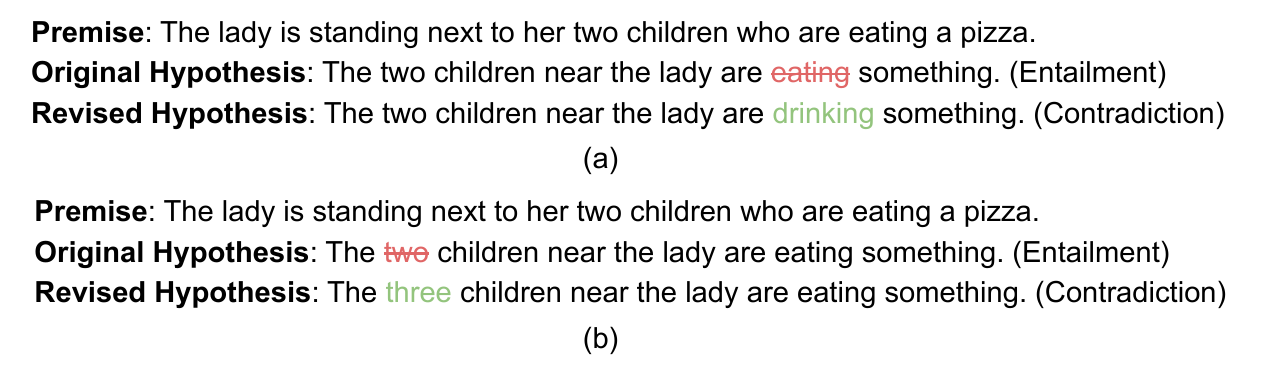}
    \caption{Illustration of counterfactual examples in natural language inference. Augmenting examples like (a) hurts performance on examples like (b) where a different robust feature has been perturbed, since the first example encourages the model to exclusively focus on the highlighted words.}
    \label{fig:eg}
\end{figure*}

A recent promising direction is to collect counterfactually-augmented data (CAD) by asking humans to minimally edit examples to flip their ground-truth label~\citep{kaushik2019learning}.
Figure \ref{fig:eg} shows example edits for Natural Language Inference (NLI).
Given interventions on \emph{robust features} that ``cause'' the label to change,
the model is expected to learn to disentangle the spurious and robust features.

Despite recent attempt to explain the efficacy of CAD by analyzing the underlying causal structure of the data~\citep{kaushik2021explaining},
empirical results on out-of-distribution (OOD) generalization using CAD are mixed.
Specifically, \citet{huang-etal-2020-counterfactually} show that CAD does not improve OOD  generalization for NLI; 
\citet{khashabi-etal-2020-bang} find that for question answering, CAD is helpful only when it is much cheaper to create than standard examples --- but ~\citet{bowman-etal-2020-new} report that the cost is actually similar per example.

In this work, we take a step towards bridging this gap between what theory suggests and what we observe in practice in regards to CAD.
An intuitive example to illustrate our key observation is shown in Figure~\ref{fig:eg} (a), where the verb `eating' is changed to `drinking' to flip the label.
While there are many other words that could have been changed to flip the label, given only these two examples, the model learns to use \emph{only} the verbs (e.g.\ using a Naive Bayes model, all other words would have zero weights).
 As a result, this model would fail when evaluated on examples such as those in (b) where the quantifier `two' is changed to `three',
 while a model trained on the unaugmented data may learn to use the quantifiers.
 
First, we use a toy theoretical setting to formalize counterfactual augmentation, and demonstrate that with CAD, the model can learn to ignore the spurious features without explicitly intervening on them. However, we find that without perturbing all robust features to generate CAD, perturbations of one robust feature can prevent the model from learning other unperturbed robust features. Motivated by this, we set up an empirical analysis on two crowdsourced CAD datasets collected for NLI and Question Answering (QA). In the empirical analysis, we identify the robust features by categorizing the edits into different \emph{perturbation types}  \citep{wu2021polyjuice} (e.g. negating a sentence or changing the quantifiers), and show that models do not generalize well to unseen perturbation types,
sometimes even performing worse than models trained on unaugmented data.

Our analysis of the relation between perturbation types and generalization can help explain other observations such as CAD being more beneficial in the low-data regime. With increasing data size, improvement from using CAD plateaus compared to unaugmented data,
suggesting that the number of perturbation types in existing CAD datasets does not keep increasing. 

Another consequence of the lack of diversity in edits is annotation artifacts, which may produce spurious correlations similar to what happens in standard crowdsourcing procedures.
While CAD is intended to debias the dataset, surprisingly,
we find that crowdsourced CAD for NLI exacerbates word overlap bias \cite{mccoy-etal-2019-right} and negation bias \cite{gururangan2018annotation} observed in existing benchmarks.

In sum, we show that while CAD can help the model ignore spurious feature, its effectiveness in current CAD datasets is limited by the set of robust features that are perturbed. Furthermore, CAD may exacerbate spurious correlations in existing benchmarks.
Our results highlight the importance of increasing the diversity of counterfactual perturbations during crowdsourcing:
We need to elicit more diverse edits of examples that make models more robust to the complexity of language.

\section{Toy Example: Analysis of a Linear Model}
\label{sec:analysis}
In this section, we use a toy setting with a linear Gaussian model and squared loss to formalize counterfactual augmentation and discuss the conditions required for it's effectiveness. The toy example serves to motivate our empirical analysis in Section~\ref{sec:robust}.

\subsection{Learning under Spurious Correlation}
We adopt the setting in \citet{rosenfeld2021the}:
each example consists of \emph{robust features} $x_r\in\BR^{d_r}$
whose joint distribution with the label is invariant during training and testing,
and \emph{spurious features} $x_s\in\BR^{d_s}$ whose joint distribution with the label varies at test time.
Here $d_r$ and $d_s$ denote the feature dimensions.
We consider a binary classification setting where the label $y \in \{-1,1\}$ is drawn from a uniform distribution, and
both the robust and spurious features are drawn from Gaussian distributions.
Specifically, an example $x=[x_r, x_s]\in\BR^{d}$ is generated by the following process (where $d = d_r + d_s$):
    \begin{align}
    y &= \begin{cases}
    1 & \text{w.p.} \; 0.5 \\
    -1 & \text{otherwise}
    \end{cases} \\
        x_r &\mid y \sim \sN(y\mu_r, \sigma_r^2I) \;, & \\
        x_s &\mid y \sim \sN(y\mu_s, \sigma_s^2I) \;,
        \label{eqn:data-gen}
    \end{align}
where $\mu_r \in\BR^{d_r}$; $\mu_s\in\BR^{d_s}$;
$\sigma_r, \sigma_s \in \BR$;
and $I$ is the identity matrix.\footnote{This model corresponds to the anti-causal setting ~\citep{scholkopf2012causal}, i.e. the label causing the features.
We adopt this setting since it is consistent with how most data is generated in tasks like NLI, sentiment analysis etc.}
The corresponding data distribution is denoted by $\sD$.
Note that the relation between $y$ and the spurious features $x_s$ depends on $\mu_s$ and $\sigma_s$,
which may change at test time,
thus relying on $x_s$ may lead to poor OOD performance.

Intuitively, in this toy setting, a model trained with only access to examples from $\sD$ would not be able to differentiate between the spurious and robust features, since they play a similar role in the data generating process for $\sD$.
Formally, consider the setting with infinite samples from $\sD$ where we learn a linear model ($y=w^Tx$ where $w\in\BR^d$) by least square regression. Let $\hat{w}\in\BR^d$ be the optimal solution on $\sD$ (without any counterfactual augmentation). The closed form solution is:
\begin{align}
\label{eqn:orig_model}
    \mathrm{Cov}(x,x)\hat{w} &= \mathrm{Cov}(x,y) \nonumber\\
    \hat{w} &= \mathrm{Cov}(x,x)^{-1}\mu
\end{align}
where $\mu=[\mu_r, \mu_s] \in \BR^{d}$ and $\mathrm{Cov}(\cdot)$ denotes the covariance matrix:
\begin{align}
\label{eqn:cov}
    \mathrm{Cov}(x,x) = \begin{bmatrix}
\Sigma_r  & \mu_r\mu_s^T\\
\mu_s\mu_r^T & \Sigma_s 
\end{bmatrix},
\end{align}
where $\Sigma_r, \Sigma_s$ are covariance matrices of $x_r$ and $x_s$ respectively.
This model relies on $x_s$ whose relationship with the label $y$ can vary at test time,
thus it may have poor performance under distribution shift.
A robust model $w_{\text{inv}}$ that is invariant to spurious correlations would ignore $x_s$:
\begin{align}
\label{eqn:inv}
    {w}_{\text{inv}} = \pb{\Sigma_r^{-1}\mu_r, 0}.
\end{align}

\subsection{Counterfactual Augmentation}

The counterfactual data is generated by editing an example to flip its label.
We model the perturbation by an \emph{edit vector} $z$ that translates $x$ to change its label from $y$ to $-y$ (i.e. from 1 to -1 or vice versa).
For instance, the counterfactual example of a positive example $(x,+1)$ is $(x+z, -1)$.
Specifically, we define the edit vector to be
$z=[yz_r, yz_s]\in\BR^{d}$,
where $z_r\in\BR^{d_r}$ and $z_s\in\BR^{d_s}$ are the displacements for the robust and spurious features.
Here, $z$ is label-dependent so that examples with different labels are translated in opposite directions.
Therefore, the counterfactual example $(x^c, -y)$ generated from $(x,y)$ has the following distribution:
\begin{align}
    x_r^c &\mid -y \sim \sN(y(\mu_r+z_r), \sigma_r^2I)\;, &\\
    x_s^c &\mid -y \sim \sN(y(\mu_s+z_s), \sigma_s^2I) \;.
\end{align}
The model is then trained on the combined set of original examples $x$ and counterfactual examples $x^c$,
whose distribution is denoted by $\sD_c$.

\paragraph{Optimal edits.}
Ideally, the counterfactual data should de-correlate $x_s$ and $y$,
thus it should only perturb the robust features $x_r$, i.e. $z=[yz_r, 0]$.
To find the displacement $z_r$ that moves $x$ across the decision boundary,
we maximize the log-likelihood of the flipped label under the data generating distribution $\sD$:
\begin{align}
    z_r^* &= \argmax_{z_r\in\BR^{d_r}} \BE_{(x,y)\sim\sD} \log p(-y\mid x+[yz_r, 0]) \nonumber\\
    &= -2\mu_r .
\end{align}
Intuitively, it moves the examples towards the mean of the opposite class along coordinates of the robust features.

Using the edits specified above, if the model trained on $\sD_c$ has optimal solution $\hat{w_c}$, we have:
\begin{align}
    \mathrm{Cov}(x,x)\hat{w_c} &= \mathrm{Cov}(x,y) \nonumber\\
    \hat{w_c} &= \pb{\Sigma_r^{-1}\mu_r, 0} = w_{\text{inv}} .
\end{align}
Thus, the optimal edits recover the robust model $w_{\text{inv}}$,
demonstrating the effectiveness of CAD.

\paragraph{Incomplete edits.}
There is an important assumption made in the above result:
we have assumed \emph{all} robust features are edited.
Suppose we have two sets of robust features $x_{r1}$ and $x_{r2}$,\footnote{We assume they are conditionally independent given the label.}
then \emph{not} editing $x_{r2}$ would effectively make it appear spurious to the model and indistinguishable from $x_s$.

In practice, this happens when there are multiple robust features but only a few are perturbed during counterfactual augmentation 
(which can be common during data collection if workers rely on simple patterns to make the minimal edits).
Considering the NLI example, if all entailment examples are flipped to non-entailment ones by inserting a negation word, then the model will only rely on negation to make predictions.

\begin{table*}[t!]
{\footnotesize
    \centering
    \begin{tabular}{lL{6.3cm}L{4.8cm}R{1.5cm}}
    \toprule
       Type  & Definition & Example & \# examples (NLI/BoolQ)\\
       \midrule
      \ctrltag{negation} & \emph{Change in negation modifier} & A dog is \textcolor{cadd}{\underline{not}} fetching anything. & 200/683\\
     \ctrltag{quantifier} & \emph{Change in words with numeral POS tags} &The lady has \textcolor{cdel}{\st{many}} $\rightarrow$ \textcolor{cadd}{\underline{three}} children. &  344/414\\
      \ctrltag{lexical} & \emph{Replace few words without changing the POS tags} &The boy is \textcolor{cdel}{\st{swimming}} $\rightarrow$ \textcolor{cadd}{\underline{running}}. & 1568/1737 \\
      \ctrltag{insert} & \emph{Only insert words or short phrases} & The \textcolor{cadd}{\underline{tall}} man is digging the ground. & 1462/536\\
     \ctrltag{delete} & \emph{Only delete words or short phrases} & The \textcolor{cdel}{\st{lazy}} person just woke up. & 562/44\\
      \multirow{2}{*}{\ctrltag{resemantic}} & \emph{Replace short phrases without affecting rest of the parsing tree} 
      & The actor \textcolor{cdel}{\st{saw}} $\rightarrow$ \textcolor{cadd}{\underline{had just met}} the director. & 2760/1866\\
      \bottomrule
    \end{tabular}
    \caption{Definition of the perturbation types and the corresponding number of examples in the NLI CAD dataset released by \cite{kaushik2019learning} and the BoolQ CAD dataset released by \citet{khashabi-etal-2020-bang}. In the example edits, the deleted words are shown in \textcolor{cdel}{\st{red}} and the newly added words are shown in \textcolor{cadd}{\underline{green}}.}
    \label{tab:code_def}
    }
\end{table*}

More formally, consider the case where the original examples $x=[x_{r1}, x_{r2}, x_s]$ and counterfactual examples are generated by incomplete edits $z=[z_{r1}, 0, 0]$ that perturb only $x_{r1}$.
Using the same analysis above where $z_{r1}$ is chosen by maximum likelihood estimation, let
the model learned on the incompletely augmented data be denoted by $\hat{w}_{\text{inc}}$. We can then show that the error of the model trained from incomplete edits can be more than that of the model trained without any counterfactual augmentation under certain conditions. More formally, we have the following: 

\begin{proposition}
\label{prop:main}
Define the error for a model as $\ell(w) = \BE_{x\sim\sF}\pb{(w_{\text{inv}}^Tx - w^Tx)^2}$ where the distribution $\sF$ is the test distribution in which $x_r$ and $x_s$ are independent: $x_r \mid y \sim \sN(y\mu_r, \sigma_r^2I)$ and $x_s \sim \sN(0, I)$.

Assuming all variables have unit variance (i.e. $\sigma_r = 1$ and $\sigma_s$ = 1), $\|\mu_{r}\|$ = 1, and $\|\mu_{s}\| = 1$, we get $\ell(\hat{w}_{\text{inc}}) > \ell(\hat{w})$ if $\|\mu_{r1}\|^2 < \frac{1 + \sqrt{13}}{6} \approx 0.767$, where $\|\cdot\|$ denotes the Euclidean norm, and $\mu_{r1}$ is the mean of the perturbed robust feature $r_1$.
\end{proposition}

Intuitively, this statement says that if the norm of the edited robust features (in the incomplete-edits model) is sufficiently small, then the test error for a model with counterfactual augmentation will be more than a model trained with no augmentation.

\begin{proof}[Proof Sketch]
The proof mainly follows from algebra and using the fact that $\mathrm{Cov}(x,x)^{-1}$ is a block matrix consisting of rank-one perturbations of the identity matrix. We refer the reader to Appendix ~\ref{app:proof} for the detailed proof.
\end{proof}

Thus, Proposition~\ref{prop:main} implies that perturbing only a small subset of robust features could perform worse than no augmentation, indicating the importance of diversity in CAD. Next, we show that the problem of incomplete edits is exhibited in real CAD too.

\section{Diversity and Generalization in CAD}
\label{sec:robust}

In this section, we test the following hypothesis based on the above analysis: models trained on CAD are limited to the specific robust features that are perturbed and may not learn other unperturbed robust features.
We empirically analyze how augmenting counterfactual examples by perturbing one robust feature affects the performance on examples generated by perturbing other robust features.

\subsection{Experiment Design}

\paragraph{Perturbation types.}

Unlike the toy example, in NLU it is not easy to define robust features since they typically correspond to the semantics of the text (e.g. sentiment). Following ~\citet{kaushik2021explaining} and similar to our toy model, we define robust features as spans of text whose distribution with the label remains invariant, whereas spans of text whose dependence on the label can change during evaluation are defined as spurious features. We then use linguistically-inspired rules~\citep{wu2021polyjuice} to categorize the robust features into several \emph{perturbation types}:
\ctrltag{negation}, \ctrltag{quantifier}, \ctrltag{lexical}, \ctrltag{insert}, \ctrltag{delete} and \ctrltag{resemantic}. Table \ref{tab:code_def} gives the definitions of each type.

\paragraph{Train/test sets.}
Both the training sets and the test sets contain counterfactual examples generated by a particular perturbation type. To test the generalization from one perturbation type to another, we use two  types of test sets: \emph{aligned test sets} where examples are generated by the same perturbation type as the training data; and \emph{unaligned test sets} where examples are generated by unseen perturbation types (e.g. training on examples from \ctrltag{lexical} and testing on \ctrltag{negation}).

\begin{table*}[t!]
\begin{small}
\centering
\begin{tabular}[t]{cccccccc}
\toprule
Train Data & \ctrltag{lexical} & \ctrltag{insert} & \ctrltag{resemantic} &  \ctrltag{quantifier} & \ctrltag{negation} &   \ctrltag{delete}\\
\midrule
SNLI seed  & 75.16\textsubscript{0.32} & 74.94\textsubscript{1.05} & \textbf{76.77}\textsubscript{0.74} & \textbf{74.36}\textsubscript{0.21} & \textbf{69.25}\textsubscript{2.09} &   65.76\textsubscript{2.34} \\
\ctrltag{lexical}  & \cellcolor{lightapricot}\textbf{79.70}\textsubscript{2.07} & 68.61\textsubscript{5.26}  & 71.46\textsubscript{3.07} & 69.90\textsubscript{3.83} &	66.00\textsubscript{2.99} &   61.76\textsubscript{5.27}\\
\ctrltag{insert} & 67.83\textsubscript{3.96} & \cellcolor{lightapricot}\textbf{79.30}\textsubscript{0.39}  & 70.53\textsubscript{2.19} & 66.31\textsubscript{3.10} & 55.00\textsubscript{4.10} &   69.75\textsubscript{2.43}\\
\ctrltag{resemantic} & 77.14\textsubscript{2.12} & 76.43\textsubscript{1.05} & \cellcolor{lightapricot}75.31\textsubscript{1.10} & 71.26\textsubscript{0.36} & 66.75\textsubscript{1.69} &   \textbf{70.16}\textsubscript{1.09}\\

\bottomrule
\end{tabular}
    \caption{Accuracy of NLI CAD on both \colorbox{lightapricot}{aligned} and unaligned test sets.
    We report the mean and standard deviation across 5 random seeds. Each dataset has a total of 1400 examples. On average models perform worse on {unaligned test sets} (i.e. unseen perturbation types).}
\label{tab:nli_pert_results}
\end{small}
\end{table*}

\begin{table*}[t!]
\begin{small}
\centering
\begin{tabular}[t]{cccccccc}
\toprule
Train Data & \ctrltag{lexical} & \ctrltag{negation} & \ctrltag{resemantic} &  \ctrltag{quantifier} & \ctrltag{insert}\\
\midrule
BoolQ seed & 65.79\textsubscript{2.11} & 62.61\textsubscript{2.65} & 68.97\textsubscript{1.83} & 61.00\textsubscript{1.65} & 57.11\textsubscript{0.67} \\
\ctrltag{lexical}  & \cellcolor{lightapricot}\textbf{77.38}\textsubscript{1.04} & 64.32\textsubscript{2.18}  & \textbf{80.78}\textsubscript{1.46} & \textbf{70.75}\textsubscript{2.03} & \textbf{66.77}\textsubscript{1.35}\\
\ctrltag{negation}  & 63.18\textsubscript{1.46} & \cellcolor{lightapricot}\textbf{72.91}\textsubscript{2.31}  & 66.74\textsubscript{2.22} & 61.75\textsubscript{2.44} & 65.42\textsubscript{1.45}\\
\ctrltag{resemantic} & 72.29\textsubscript{0.72} & 64.92\textsubscript{1.56} & \cellcolor{lightapricot}75.60\textsubscript{2.11} & 70.00\textsubscript{2.85} & 64.91\textsubscript{2.31}\\
\bottomrule
\end{tabular}
    \caption{Accuracy of BoolQ CAD on both \colorbox{lightapricot}{aligned} and unaligned test sets.
    We report the mean and standard deviation across 5 random seeds. Each dataset has a total of 9427 examples. On average models perform worse on {unaligned test sets} (i.e. unseen perturbation types).}
\label{tab:qa_pert_results}
\end{small}
\end{table*}

\subsection{Experimental Setup}
\label{ssec:experiment}

\paragraph{Data.}
We experiment on two CAD datasets collected for SNLI ~\cite{kaushik2019learning} and BoolQ ~\cite{khashabi-etal-2020-bang}. The size of the paired data (seed examples and edited examples) for each perturbation type in the training dataset is given in Table~\ref{tab:code_def}. Since some types (e.g. \ctrltag{delete}) contain too few examples for training, we train on the top three largest perturbation types: \ctrltag{lexical}, \ctrltag{insert}, and \ctrltag{resemantic} for SNLI; and \ctrltag{lexical}, \ctrltag{negation}, and \ctrltag{resemantic} for BoolQ. 

For SNLI, to control for dataset sizes across all experiments, we use 700 seed examples and their corresponding 700 perturbations for each perturbation type. As a baseline (`SNLI seed'), we subsample examples from SNLI to create a similar sized dataset for comparison.\footnote{We observe similar trends when using different subsets of the SNLI data. We report the mean and standard deviation across different subsets in Appendix~\ref{app:more_expt}.}

For BoolQ~\cite{clark-etal-2019-boolq}, our initial experiments show that training on only CAD does not reach above random-guessing.
Thus, we include all original training examples in BoolQ~\cite{khashabi-etal-2020-bang}, and replace part of them with CAD for each perturbation type.
This results in a training set of 9427 examples of which 683 are CAD for each perturbation type.
The size 683 is chosen to match the the smallest CAD type for BoolQ.
As a baseline (`BoolQ seed'), we train on all the original training examples, consisting of 9427 examples. For both datasets, the training, dev and test sets are created from their respective splits in the CAD datasets. The size of the dev and test sets is reported in Appendix ~\ref{app:data_details}.

\paragraph{Model.}
We use the Hugging Face implementation \citep{Wolf2019HuggingFacesTS} of RoBERTa \citep{Liu2019RoBERTaAR} to fine-tune all our models.
To account for the small dataset sizes, we run all our experiments with 5 different random seeds and report the mean and standard deviation. Details on hyperparameter tuning are reported in Appendix ~\ref{app:exp_details}.\footnote{Our code can be found at: \url{https://github.com/joshinh/investigation-cad}}

\subsection{Generalization to Unseen Perturbation Types}
\label{sec:results_perturb_types}

We discuss results for the main question in this section---how does adding CAD generated from one perturbation type affect performance on examples generated from other perturbation types? Table~\ref{tab:nli_pert_results} and ~\ref{tab:qa_pert_results} show results for SNLI and BoolQ.

\paragraph{CAD performs well on aligned test sets.}
We see that on average models perform very well on the \colorbox{lightapricot}{aligned test sets} (same perturbation type as the training set), but do not always do well on unaligned test sets (unseen perturbation types), which is consistent with our analysis in Section~\ref{sec:analysis}.
On SNLI, one exception is \ctrltag{resemantic},
which performs well on unseen perturbation types.
We believe this is because it is a broad category (replacing any constituent) that covers other types such as \ctrltag{lexical} (replacing any word). Similarly, on BoolQ, \ctrltag{lexical} and \ctrltag{resemantic} both perform better than the baseline on some unaligned test sets (e.g. \ctrltag{quantifier}), but they perform much better on the aligned test sets.

\begin{figure}[ht!]
    \centering
    \includegraphics[scale=0.49]{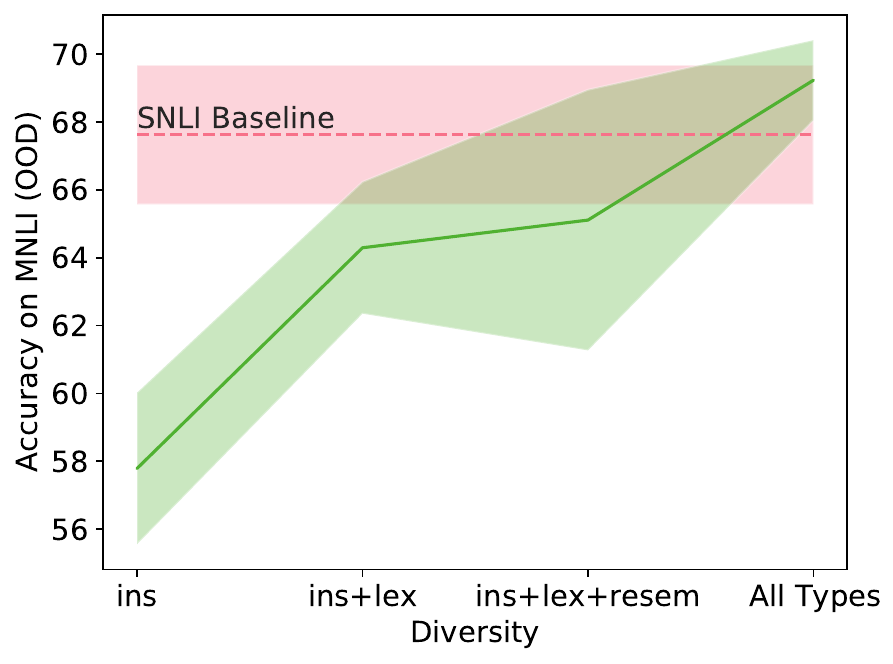}
    \caption{OOD accuracy (mean, std. deviation) on MNLI of models trained on SNLI CAD and SNLI seed (baseline) with increasing number of perturbation types and fixed training set size. More perturbation types in the training data leads to higher OOD accuracy.}
    \label{fig:incr_diversity}
\end{figure}

\begin{figure*}[ht!]
\centering
    \begin{minipage}[t]{0.49\linewidth}
    \centering
    \includegraphics[width=0.99\linewidth]{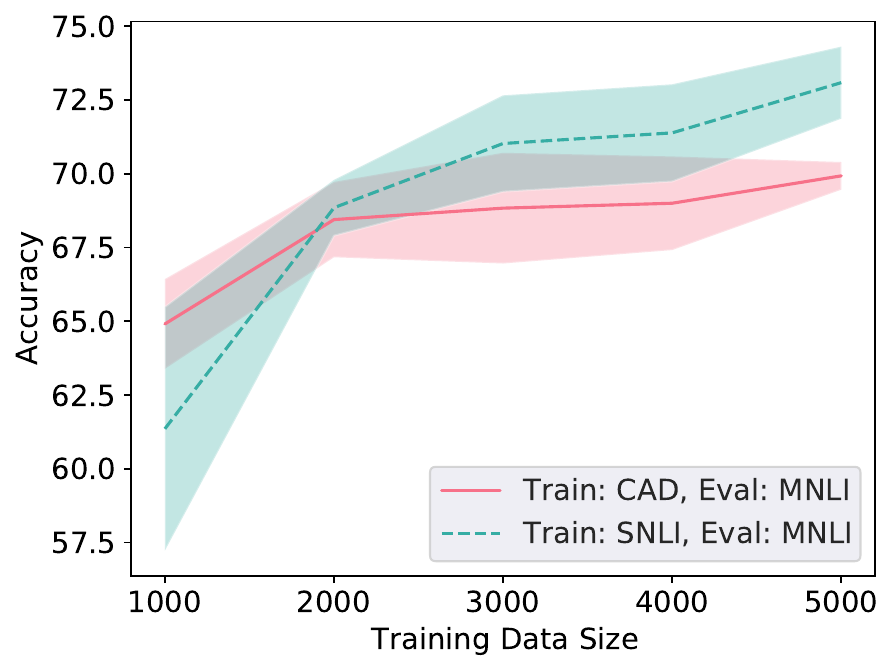}
    \vspace{-1pt}
    \caption{Accuracy on the OOD set (MNLI) for models trained on increasing amounts of NLI CAD. CAD is more beneficial in the low data regime, but its benefits taper off (compared to SNLI baseline of same size) as the dataset size increases.}
    \label{fig:nli_incr_data}
    \end{minipage}
    \hfill
    \begin{minipage}[t]{0.49\linewidth}
    \centering
    \includegraphics[width=0.99\linewidth]{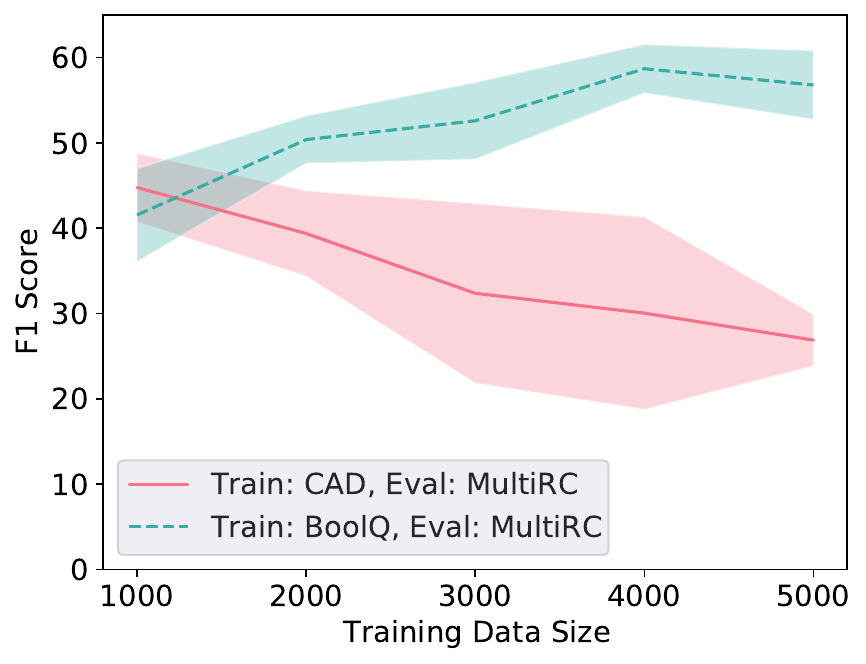}
    \vspace{-1pt}
    \caption{F1 score on the OOD set (MultiRC) for models trained on increasing amounts of QA CAD. CAD performs comparable to the baseline in the low data regime, but surprisingly performs worse with increasing dataset sizes, probably due to overfitting to a few perturbation types.}
    \label{fig:qa_incr_data}
    \end{minipage}
\end{figure*}

\paragraph{CAD sometimes performs worse than the baseline on unaligned test sets.}
For example, on SNLI, training on \ctrltag{insert} does much worse than the seed baseline on  \ctrltag{lexical} and \ctrltag{resemantic}, and SNLI seed performs best on \ctrltag{quantifier} and \ctrltag{negation}. On BoolQ, training on \ctrltag{negation} does slightly worse than the baseline on \ctrltag{lexical} and \ctrltag{resemantic}. This suggests that augmenting perturbations of one particular robust feature may reduce the model's reliance on other robust features, that could have been learned without augmentation.

\subsection{Generalization to Out-of-Distribution Data}

In \refsec{results_perturb_types}, we have seen that training on CAD generated by a single perturbation type does not generalize well to unseen perturbation types.
However, in practice CAD contains many different perturbation types.
Do they cover enough robust features to enable OOD generalization?

\paragraph{Increasing Diversity.} We first verify that increasing the number of perturbed robust features leads to better OOD generalization. Specifically, we train models on subsets of SNLI CAD with increasing coverage of perturbation types and evaluate on MNLI as the OOD data.
Starting with only \ctrltag{insert},
we add one perturbation type at a time until all types are included;
the total number of examples are fixed throughout the process at 1400 (which includes 700 seed examples and the corresponding 700 perturbations).

Figure~\ref{fig:incr_diversity} shows the OOD accuracy on MNLI when trained on CAD and SNLI seed examples of the same size. We observe that as the number of perturbation types increases, models generalize better to OOD data despite fixed training data size. The result highlights the importance of collecting a diverse set of counterfactual examples, even if each perturbation type is present in a small amount.

A natural question to ask here is: If we continue to collect more counterfactual data, does it cover more perturbation types and hence lead to better OOD generalization? Thus we investigate the impact of training data size next.\footnote{The results in Figure~\ref{fig:incr_diversity} when all perturbation types are included indicate that CAD performs better than the SNLI baseline. This is not in contradiction with the results found in \citet{huang-etal-2020-counterfactually}, since our models are trained on only a subset of CAD. This further motivates the study of how CAD data size affects generalization.}

\paragraph{Role of Dataset Size.} To better understand the role dataset size plays in OOD generalization, we plot the learning curve on SNLI CAD in Figure~\ref{fig:nli_incr_data}, where we gradually increase the amount of CAD for training. The baseline model is trained on SNLI seed examples of the same size, and all models are evaluated on MNLI (as the OOD dataset). We also conduct a similar experiment on BoolQ in Figure~\ref{fig:qa_incr_data},
where a subset of MultiRC~\cite{MultiRC2018} is used as the OOD dataset following~\citet{khashabi-etal-2020-bang}.
Since the test set is unbalanced, we report F1 scores instead of accuracy in this case.

\begin{table}[t]
\begin{small}
    \centering
    \begin{tabular}{ccc}
    \toprule
         & BERT & RoBERTa\\
    \midrule
    SNLI seed & 59.7\textsubscript{0.3}  & \textbf{73.8}\textsubscript{1.2} \\
    CAD & \textbf{60.2}\textsubscript{1.0} & 70.0\textsubscript{1.1}\\
    \bottomrule
    \end{tabular}
    \caption{Accuracy (mean and std. deviation across 5 runs) on MNLI of different pretrained models fine-tuned on SNLI seed and CAD. CAD seems to be less beneficial when using better pretrained models.}
    \label{tab:nli_pretraining}
\end{small}
\end{table}

For SNLI, CAD is beneficial for OOD generalization only in low data settings ($<$ 2000 examples). As the amount of data increases, the comparable SNLI baseline performs  better and surpasses the performance of CAD. Similarly for BoolQ, we observe that CAD is comparable to the baseline in the low data setting ($\sim$ 1000 examples). Surprisingly, more CAD for BoolQ leads to worse OOD performance. We suspect this is due to overfitting to the specific perturbation types present in BoolQ CAD.

Intuitively, as we increase the amount of data, the diversity of robust features covered by the seed examples also increases. On the other hand, the benefit of CAD is restricted to the \emph{perturbed} robust features. 
The plateaued performance of CAD (in the case of NLI) shows that the diversity of perturbations may not increase with the data size as fast as we would like,
calling for better crowdsourcing protocols to elicit diverse edits from workers.

\paragraph{Role of Pretraining.} 
\citet{tu-etal-2020-empirical} show that larger pretrained models generalize better from minority examples.
Therefore, in our case we would expect CAD to have limited benefit on larger pretrained models since they can already leverage the diverse (but scarce) robust features revealed by SNLI examples. We compare the results of BERT \cite{devlin-etal-2019-bert} and RoBERTa \cite{Liu2019RoBERTaAR} trained on  SNLI CAD in Table \ref{tab:nli_pretraining} --- both models are fine-tuned on the SNLI CAD dataset and are evaluated on the OOD set (MNLI). For the RoBERTa model (pretrained on more data), CAD no longer improves over the SNLI baseline,
suggesting that current CAD datasets may not have much better coverage of robust features than what stronger pretrained models can already learn from benchmarks like SNLI.

\begin{figure*}[t!]
\centering
\begin{subfigure}{0.49\textwidth}
    \centering
    \includegraphics[scale=0.49]{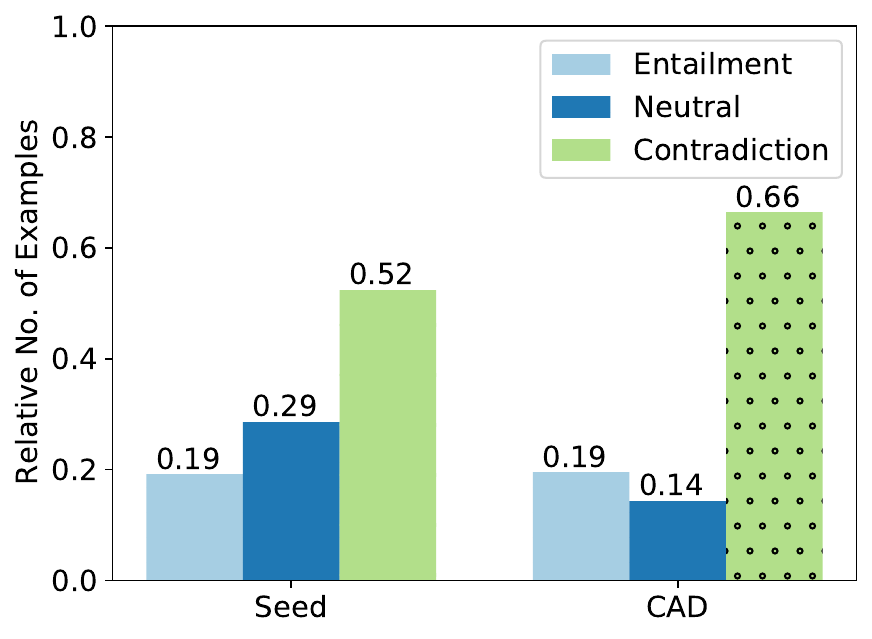}
    \vspace{-1pt}
    \caption{Negation bias}
    \label{fig:negation_bias}
\end{subfigure}
\begin{subfigure}{0.49\textwidth}
    \centering
    \includegraphics[scale=0.49]{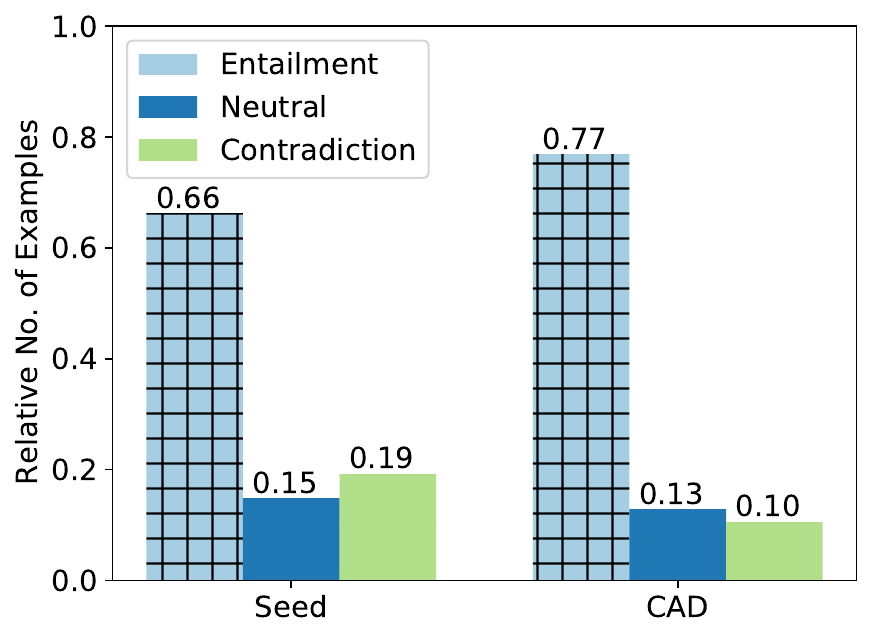}
    \vspace{-1pt}
    \caption{Word-overlap bias}
    \label{fig:word_overlap_bias}
\end{subfigure}
\vspace{-1pt}
\caption{Fraction of entailment/neutral/contradiction examples in the SNLI seed set and CAD where (a) negation words are present in the hypothesis; (b) word overlap bias is observed. We observe that the distribution is more skewed in CAD compared to the seed examples, towards contradiction for the negation bias (a) and towards entailment for the word overlap bias (b).}
\label{}
\end{figure*}

\section{CAD Exacerbates Existing Spurious Correlation}
\label{sec:spurious}

An artifact of underdiverse perturbations is the newly introduced spurious correlations. As an example, in the extreme case where all entailment examples are flipped to non-entailment by the \ctrltag{negation} operation in Table~\ref{tab:code_def}, the model would learn to exclusively rely on the existence of negation words to make predictions,
which is clearly undesirable.
In this section, we study the impact of CAD on two known spurious correlations in NLI benchmarks: word overlap bias \cite{mccoy-etal-2019-right} and negation bias \cite{Gururangan2018AnnotationAI}.

\begin{table}[t]
\begin{small}
    \centering
    \begin{tabular}{ccc}
    \toprule
         & {Stress Test} & {MNLI subset} \\
    \midrule
        SNLI Seed & \textbf{57.5}\textsubscript{4.6} & \textbf{63.3}\textsubscript{3.8} \\
        CAD & 49.6\textsubscript{1.5} & 55.7\textsubscript{4.2}\\
    \bottomrule
    \end{tabular}
    \vspace{-2pt}
    \caption{Accuracy of models on challenge examples in the stress test and MNLI, where non-contradiction examples contain a negation word in the hypothesis. Models trained on CAD perform worse on both sets, implying that it exacerbates the negation bias.}
    \label{tab:negation}
\end{small}
\end{table}

\paragraph{Negation bias.} We take examples where there is a presence of a negation word (i.e. "no", "not", "n't") in the hypothesis, and plot the fraction of examples in each class in both the seed and the corresponding CAD examples in Figure~\ref{fig:negation_bias}.
As expected, contradiction is the majority class in the seed group,
but surprisingly, 
including CAD amplifies the fraction of contradiction examples! 
As a result, training on CAD leads to worse performance on challenge sets that counter the negation bias compared to training on seed examples of the same size.
Specifically, we test on the `negation' part of the Stress Tests \cite{naik-etal-2018-stress}\footnote{Synthetic examples where the phrase ``and false is not true'' is appended to the hypothesis of MNLI examples.}
and challenge examples in the combined MNLI development set which contain negation words in the hypothesis but are not contradictions. 
Table~\ref{tab:negation} shows that models trained on CAD perform worse on both test sets, implying that they rely more on the negation bias.

\paragraph{Word-overlap bias.}
Similarly, in \reffig{word_overlap_bias}, we show that CAD amplifies the fraction of entailment examples among those with high word overlap (i.e.\ more than 90\% of words in the hypothesis are present in the premise).
Models trained on SNLI and CAD both perform poorly (< 10\% accuracy) on the non-entailment subset of HANS challenge set \cite{mccoy-etal-2019-right}, which exploits the word overlap bias. 

\paragraph{Takeaway.} This section reveals that in the process of creating CAD, we may inadvertently exacerbate existing spurious correlations. The fundamental challenge here is that perturbations of the robust features are only observed through word change in the sentence---it is hard to surface the underlying causal variables without introducing (additional) artifacts to the sentence form.

\section{Related Work}
\label{sec:related}

\paragraph{Label-Preserving Data Augmentation.} A common strategy to build more robust models is to augment existing datasets with examples similar to those from the target distribution. \citet{min2020augmentation} improve accuracy on HANS challenge set~\citep{mccoy-etal-2019-right} by augmenting syntactically-rich examples. \citet{jia-liang-2016-data} and \citet{Andreas2020GoodEnoughCD} recombine examples to achieve better compositional generalization.
There has also been a recent body of work using task-agnostic data augmentation
by paraphrasing \cite{Wei2019EDAED},
back-translation \cite{sennrich-etal-2016-improving}  
and masked language models \cite{ng-etal-2020-ssmba}. The main difference between these works and CAD is that the edits in these works are label-preserving whereas they are label-flipping in CAD---the former prevents models from being over-sensitive and the latter alleviates under-sensitivity to perturbations. 

\paragraph{Label-Changing Data Augmentation.} \citet{Lu2020GenderBI} and \citet{Zmigrod2019CounterfactualDA} use rule-based CAD to mitigate gender stereotypes. \citet{gardner2020evaluating} build similar contrast sets using expert edits for evaluation. In contrast, \citet{kaushik2019learning} crowdsource minimal edits. Recently, \citet{damien2020learning} also use CAD along with additional auxiliary training objectives and demonstrate improved OOD generalization.

\citet{kaushik2021explaining} analyze a similar toy model (linear Gaussian model) demonstrating the benefits of CAD, and showed that noising the edited spans hurts performance more than other spans. Our analysis complements theirs by showing that
while spans identified by CAD are useful, a lack of diversity in these spans limit the effectiveness of CAD, thus better coverage of robust features could potentially lead to better OOD generalization.

\paragraph{Robust Learning Algorithms.}  Another direction of work has explored learning more robust models without using additional augmented data. These methods essentially rely on learning debiased representations---\citet{wang2018learning} create a biased classifier and project its representation out of the model's representation. Along similar lines, \citet{belinkov-etal-2019-dont} remove hypothesis-only bias in NLI models by adversarial training. \citet{he2019unlearn} and \citet{Clark2019DontTT} correct the conditional distribution given a biased model. \citet{utama-etal-2020-towards} build on this to remove `unknown' biases, assuming that a weak model learns a biased representations. More recently, \citet{veitch2021counterfactual} use ideas from causality to learn invariant predictors from counterfactual examples. 
The main difference between these methods and CAD is that the former generally requires some prior knowledge of what spurious correlations models learn (e.g.\ by constructing a biased model or weak model), whereas CAD is a more general human-in-the-loop method that leverages humans' knowledge of robust features.

\section{Conclusion and Future Directions}
\label{sec:conclusion}

In this work, we first analyzed CAD theoretically using a linear model and showed that models do not generalize to unperturbed robust features. We then empirically demonstrated this issue in two CAD datasets, where models do not generalize well to unseen perturbation types. 
We also showed that CAD amplifies existing spurious correlations, pointing out another concern. Given these results, a natural question is: How can we fix these problems and make CAD more useful for OOD generalization? We  discuss a few directions which we think could be helpful:

\begin{itemize}
    \item We can use generative models \cite{2020t5, lewis-etal-2020-bart} to generate \emph{diverse} minimal perturbations and then crowdsource labels for them \cite{wu2021polyjuice}. We can improve the diversity of the generations by masking different spans in the text to be in-filled, thus covering more robust features.
    \item An alternative to improving the crowdsourcing procedure is to devise better learning algorithms which mitigate the issues pointed out in this work. For example, given that we know the models do not always generalize well to unperturbed features, we can regularize the model to limit the reliance on the perturbed features.
\end{itemize}

We hope that this analysis spurs future work on CAD, making them more useful for OOD generalization.

\section*{Acknowledgements}

We thank Divyansh Kaushik, Tatsunori Hashimoto and members of the NYU ML2 group for discussion and feedback on the work. The first author is supported by a NSF Graduate Research Fellowship under grant number 1839302. This work was partly supported by Samsung Advanced Institute of Technology (Next Generation Deep Learning: From Pattern Recognition to AI).

\bibliography{tacl2018, all}

\begin{thebibliography}{41}
\expandafter\ifx\csname natexlab\endcsname\relax\def\natexlab#1{#1}\fi

\bibitem[{Andreas(2020)}]{Andreas2020GoodEnoughCD}
Jacob Andreas. 2020.
\newblock Good-enough compositional data augmentation.
\newblock In \emph{ACL}.

\bibitem[{Belinkov et~al.(2019)Belinkov, Poliak, Shieber, Van~Durme, and
  Rush}]{belinkov-etal-2019-dont}
Yonatan Belinkov, Adam Poliak, Stuart Shieber, Benjamin Van~Durme, and
  Alexander Rush. 2019.
\newblock Don{'}t take the premise for granted: Mitigating artifacts in natural
  language inference.
\newblock In \emph{Proceedings of the 57th Annual Meeting of the Association
  for Computational Linguistics}, Florence, Italy. Association for
  Computational Linguistics.

\bibitem[{Bowman et~al.(2020)Bowman, Palomaki, Baldini~Soares, and
  Pitler}]{bowman-etal-2020-new}
Samuel~R. Bowman, Jennimaria Palomaki, Livio Baldini~Soares, and Emily Pitler.
  2020.
\newblock New protocols and negative results for textual entailment data
  collection.
\newblock In \emph{Proceedings of the 2020 Conference on Empirical Methods in
  Natural Language Processing (EMNLP)}, Online. Association for Computational
  Linguistics.

\bibitem[{Clark et~al.(2019{\natexlab{a}})Clark, Lee, Chang, Kwiatkowski,
  Collins, and Toutanova}]{clark-etal-2019-boolq}
Christopher Clark, Kenton Lee, Ming-Wei Chang, Tom Kwiatkowski, Michael
  Collins, and Kristina Toutanova. 2019{\natexlab{a}}.
\newblock {B}ool{Q}: Exploring the surprising difficulty of natural yes/no
  questions.
\newblock In \emph{NAACL}.

\bibitem[{Clark et~al.(2019{\natexlab{b}})Clark, Yatskar, and
  Zettlemoyer}]{Clark2019DontTT}
Christopher Clark, Mark Yatskar, and Luke Zettlemoyer. 2019{\natexlab{b}}.
\newblock Don't take the easy way out: Ensemble based methods for avoiding
  known dataset biases.
\newblock In \emph{EMNLP/IJCNLP}.

\bibitem[{Devlin et~al.(2019)Devlin, Chang, Lee, and
  Toutanova}]{devlin-etal-2019-bert}
Jacob Devlin, Ming-Wei Chang, Kenton Lee, and Kristina Toutanova. 2019.
\newblock \href {https://doi.org/10.18653/v1/N19-1423} {{BERT}: Pre-training of
  deep bidirectional transformers for language understanding}.
\newblock In \emph{Proceedings of the 2019 Conference of the North {A}merican
  Chapter of the Association for Computational Linguistics: Human Language
  Technologies, Volume 1 (Long and Short Papers)}, pages 4171--4186,
  Minneapolis, Minnesota. Association for Computational Linguistics.

\bibitem[{Gardner et~al.(2020)Gardner, Artzi, Basmova, Berant, Bogin, Chen,
  Dasigi, Dua, Elazar, Gottumukkala, Gupta, Hajishirzi, Ilharco, Khashabi, Lin,
  Liu, Liu, Mulcaire, Ning, Singh, Smith, Subramanian, Tsarfaty, Wallace,
  Zhang, and Zhou}]{gardner2020evaluating}
M.~Gardner, Y.~Artzi, V.~Basmova, J.~Berant, B.~Bogin, S.~Chen, P.~Dasigi,
  D.~Dua, Y.~Elazar, A.~Gottumukkala, N.~Gupta, H.~Hajishirzi, G.~Ilharco,
  D.~Khashabi, K.~Lin, J.~Liu, N.~F. Liu, P.~Mulcaire, Q.~Ning, S.~Singh, N.~A.
  Smith, S.~Subramanian, R.~Tsarfaty, E.~Wallace, A.~Zhang, and B.~Zhou. 2020.
\newblock Evaluating {NLP} models via contrast sets.
\newblock In \emph{Empirical Methods in Natural Language Processing (EMNLP)}.

\bibitem[{Gururangan et~al.(2018{\natexlab{a}})Gururangan, Swayamdipta, Levy,
  Schwartz, Bowman, and Smith}]{gururangan2018annotation}
S.~Gururangan, S.~Swayamdipta, O.~Levy, R.~Schwartz, S.~R. Bowman, and N.~A.
  Smith. 2018{\natexlab{a}}.
\newblock Annotation artifacts in natural language inference data.
\newblock In \emph{North American Association for Computational Linguistics
  (NAACL)}.

\bibitem[{Gururangan et~al.(2018{\natexlab{b}})Gururangan, Swayamdipta, Levy,
  Schwartz, Bowman, and Smith}]{Gururangan2018AnnotationAI}
Suchin Gururangan, Swabha Swayamdipta, Omer Levy, Roy Schwartz, Samuel~R.
  Bowman, and Noah~A. Smith. 2018{\natexlab{b}}.
\newblock Annotation artifacts in natural language inference data.
\newblock In \emph{NAACL-HLT}.

\bibitem[{He et~al.(2019)He, Zha, and Wang}]{he2019unlearn}
H.~He, S.~Zha, and H.~Wang. 2019.
\newblock Unlearn dataset bias for natural language inference by fitting the
  residual.
\newblock In \emph{Proceedings of the EMNLP Workshop on Deep Learning for
  Low-Resource {NLP}}.

\bibitem[{Huang et~al.(2020)Huang, Liu, and
  Bowman}]{huang-etal-2020-counterfactually}
William Huang, Haokun Liu, and Samuel~R. Bowman. 2020.
\newblock \href {https://doi.org/10.18653/v1/2020.insights-1.13}
  {Counterfactually-augmented {SNLI} training data does not yield better
  generalization than unaugmented data}.
\newblock In \emph{Proceedings of the First Workshop on Insights from Negative
  Results in NLP}, pages 82--87, Online. Association for Computational
  Linguistics.

\bibitem[{Jia and Liang(2016)}]{jia-liang-2016-data}
Robin Jia and Percy Liang. 2016.
\newblock \href {https://doi.org/10.18653/v1/P16-1002} {Data recombination for
  neural semantic parsing}.
\newblock In \emph{Proceedings of the 54th Annual Meeting of the Association
  for Computational Linguistics (Volume 1: Long Papers)}, pages 12--22, Berlin,
  Germany. Association for Computational Linguistics.

\bibitem[{Kaushik et~al.(2020)Kaushik, Hovy, and Lipton}]{kaushik2019learning}
Divyansh Kaushik, Eduard Hovy, and Zachary~C Lipton. 2020.
\newblock Learning the difference that makes a difference with
  counterfactually-augmented data.
\newblock In \emph{International Conference on Learning Representations
  (ICLR)}.

\bibitem[{Kaushik et~al.(2021)Kaushik, Setlur, Hovy, and
  Lipton}]{kaushik2021explaining}
Divyansh Kaushik, Amrith Setlur, Eduard~H Hovy, and Zachary~Chase Lipton. 2021.
\newblock \href {https://openreview.net/forum?id=HHiiQKWsOcV} {Explaining the
  efficacy of counterfactually augmented data}.
\newblock In \emph{International Conference on Learning Representations}.

\bibitem[{Khashabi et~al.(2018)Khashabi, Chaturvedi, Roth, Upadhyay, and
  Roth}]{MultiRC2018}
Daniel Khashabi, Snigdha Chaturvedi, Michael Roth, Shyam Upadhyay, and Dan
  Roth. 2018.
\newblock Looking beyond the surface:a challenge set for reading comprehension
  over multiple sentences.
\newblock In \emph{NAACL}.

\bibitem[{Khashabi et~al.(2020)Khashabi, Khot, and
  Sabharwal}]{khashabi-etal-2020-bang}
Daniel Khashabi, Tushar Khot, and Ashish Sabharwal. 2020.
\newblock \href {https://doi.org/10.18653/v1/2020.emnlp-main.12} {More bang for
  your buck: Natural perturbation for robust question answering}.
\newblock In \emph{Proceedings of the 2020 Conference on Empirical Methods in
  Natural Language Processing (EMNLP)}, pages 163--170, Online. Association for
  Computational Linguistics.

\bibitem[{Lewis et~al.(2020)Lewis, Liu, Goyal, Ghazvininejad, Mohamed, Levy,
  Stoyanov, and Zettlemoyer}]{lewis-etal-2020-bart}
Mike Lewis, Yinhan Liu, Naman Goyal, Marjan Ghazvininejad, Abdelrahman Mohamed,
  Omer Levy, Veselin Stoyanov, and Luke Zettlemoyer. 2020.
\newblock \href {https://doi.org/10.18653/v1/2020.acl-main.703} {{BART}:
  Denoising sequence-to-sequence pre-training for natural language generation,
  translation, and comprehension}.
\newblock In \emph{Proceedings of the 58th Annual Meeting of the Association
  for Computational Linguistics}, pages 7871--7880, Online. Association for
  Computational Linguistics.

\bibitem[{Liu et~al.(2019)Liu, Ott, Goyal, Du, Joshi, Chen, Levy, Lewis,
  Zettlemoyer, and Stoyanov}]{Liu2019RoBERTaAR}
Y.~Liu, Myle Ott, Naman Goyal, Jingfei Du, Mandar Joshi, Danqi Chen, Omer Levy,
  M.~Lewis, Luke Zettlemoyer, and Veselin Stoyanov. 2019.
\newblock {RoBERTa}: A robustly optimized {BERT} pretraining approach.
\newblock \emph{ArXiv}, abs/1907.11692.

\bibitem[{Lu et~al.(2020)Lu, Mardziel, Wu, Amancharla, and
  Datta}]{Lu2020GenderBI}
Kaiji Lu, Piotr Mardziel, Fangjing Wu, Preetam Amancharla, and A.~Datta. 2020.
\newblock Gender bias in neural natural language processing.
\newblock In \emph{Logic, Language, and Security}.

\bibitem[{McCoy et~al.(2019)McCoy, Pavlick, and Linzen}]{mccoy-etal-2019-right}
Tom McCoy, Ellie Pavlick, and Tal Linzen. 2019.
\newblock \href {https://doi.org/10.18653/v1/P19-1334} {Right for the wrong
  reasons: Diagnosing syntactic heuristics in natural language inference}.
\newblock In \emph{Proceedings of the 57th Annual Meeting of the Association
  for Computational Linguistics}, pages 3428--3448, Florence, Italy.
  Association for Computational Linguistics.

\bibitem[{Min et~al.(2020)Min, McCoy, Das, Pitler, and
  Linzen}]{min2020augmentation}
Junghyun Min, R.~Thomas McCoy, Dipanjan Das, Emily Pitler, and Tal Linzen.
  2020.
\newblock Syntactic data augmentation increases robustness to inference
  heuristics.
\newblock In \emph{Proceedings of the 58th Annual Meeting of the Association
  for Computational Linguistics}, Seattle, Washington. Association for
  Computational Linguistics.

\bibitem[{Naik et~al.(2018)Naik, Ravichander, Sadeh, Rose, and
  Neubig}]{naik-etal-2018-stress}
Aakanksha Naik, Abhilasha Ravichander, Norman Sadeh, Carolyn Rose, and Graham
  Neubig. 2018.
\newblock \href {https://www.aclweb.org/anthology/C18-1198} {Stress test
  evaluation for natural language inference}.
\newblock In \emph{Proceedings of the 27th International Conference on
  Computational Linguistics}, pages 2340--2353, Santa Fe, New Mexico, USA.
  Association for Computational Linguistics.

\bibitem[{Ng et~al.(2020)Ng, Cho, and Ghassemi}]{ng-etal-2020-ssmba}
Nathan Ng, Kyunghyun Cho, and Marzyeh Ghassemi. 2020.
\newblock {SSMBA}: Self-supervised manifold based data augmentation for
  improving out-of-domain robustness.
\newblock In \emph{Proceedings of the 2020 Conference on Empirical Methods in
  Natural Language Processing (EMNLP)}, Online. Association for Computational
  Linguistics.

\bibitem[{Peters et~al.(2018)Peters, Neumann, Iyyer, Gardner, Clark, Lee, and
  Zettlemoyer}]{peters-etal-2018-deep}
Matthew Peters, Mark Neumann, Mohit Iyyer, Matt Gardner, Christopher Clark,
  Kenton Lee, and Luke Zettlemoyer. 2018.
\newblock \href {https://doi.org/10.18653/v1/N18-1202} {Deep contextualized
  word representations}.
\newblock In \emph{Proceedings of the 2018 Conference of the North {A}merican
  Chapter of the Association for Computational Linguistics: Human Language
  Technologies, Volume 1 (Long Papers)}, pages 2227--2237, New Orleans,
  Louisiana. Association for Computational Linguistics.

\bibitem[{Raffel et~al.(2020)Raffel, Shazeer, Roberts, Lee, Narang, Matena,
  Zhou, Li, and Liu}]{2020t5}
Colin Raffel, Noam Shazeer, Adam Roberts, Katherine Lee, Sharan Narang, Michael
  Matena, Yanqi Zhou, Wei Li, and Peter~J. Liu. 2020.
\newblock \href {http://jmlr.org/papers/v21/20-074.html} {Exploring the limits
  of transfer learning with a unified text-to-text transformer}.
\newblock \emph{Journal of Machine Learning Research}, 21(140):1--67.

\bibitem[{Rajpurkar et~al.(2016)Rajpurkar, Zhang, Lopyrev, and
  Liang}]{rajpurkar-etal-2016-squad}
Pranav Rajpurkar, Jian Zhang, Konstantin Lopyrev, and Percy Liang. 2016.
\newblock \href {https://doi.org/10.18653/v1/D16-1264} {{SQ}u{AD}: 100,000+
  questions for machine comprehension of text}.
\newblock In \emph{Proceedings of the 2016 Conference on Empirical Methods in
  Natural Language Processing}, pages 2383--2392, Austin, Texas. Association
  for Computational Linguistics.

\bibitem[{Rosenfeld et~al.(2021)Rosenfeld, Ravikumar, and
  Risteski}]{rosenfeld2021the}
Elan Rosenfeld, Pradeep~Kumar Ravikumar, and Andrej Risteski. 2021.
\newblock \href {https://openreview.net/forum?id=BbNIbVPJ-42} {The risks of
  invariant risk minimization}.
\newblock In \emph{International Conference on Learning Representations}.

\bibitem[{Scholkopf et~al.(2012)Scholkopf, Janzing, Peters, Sgouritsa, Zhang,
  and Mooij}]{scholkopf2012causal}
B.~Scholkopf, D.~Janzing, J.~Peters, E.~Sgouritsa, K.~Zhang, and J.~Mooij.
  2012.
\newblock On causal and anticausal learning.
\newblock In \emph{International Conference on Machine Learning (ICML)}.

\bibitem[{Sennrich et~al.(2016)Sennrich, Haddow, and
  Birch}]{sennrich-etal-2016-improving}
Rico Sennrich, Barry Haddow, and Alexandra Birch. 2016.
\newblock \href {https://doi.org/10.18653/v1/P16-1009} {Improving neural
  machine translation models with monolingual data}.
\newblock In \emph{Proceedings of the 54th Annual Meeting of the Association
  for Computational Linguistics (Volume 1: Long Papers)}, pages 86--96, Berlin,
  Germany. Association for Computational Linguistics.

\bibitem[{Teney et~al.(2020)Teney, Abbasnedjad, and van~den
  Hengel}]{damien2020learning}
Damien Teney, Ehsan Abbasnedjad, and Anton van~den Hengel. 2020.
\newblock Learning what makes a difference from counterfactual examples and
  gradient supervision.
\newblock In \emph{Computer Vision -- ECCV 2020}, pages 580--599, Cham.
  Springer International Publishing.

\bibitem[{Tu et~al.(2020)Tu, Lalwani, Gella, and He}]{tu-etal-2020-empirical}
Lifu Tu, Garima Lalwani, Spandana Gella, and He~He. 2020.
\newblock \href {https://doi.org/10.1162/tacl_a_00335} {An empirical study on
  robustness to spurious correlations using pre-trained language models}.
\newblock \emph{Transactions of the Association for Computational Linguistics},
  8:621--633.

\bibitem[{Utama et~al.(2020)Utama, Moosavi, and
  Gurevych}]{utama-etal-2020-towards}
Prasetya~Ajie Utama, Nafise~Sadat Moosavi, and Iryna Gurevych. 2020.
\newblock Towards debiasing {NLU} models from unknown biases.
\newblock In \emph{Proceedings of the 2020 Conference on Empirical Methods in
  Natural Language Processing (EMNLP)}, Online. Association for Computational
  Linguistics.

\bibitem[{Veitch et~al.(2021)Veitch, D'Amour, Yadlowsky, and
  Eisenstein}]{veitch2021counterfactual}
Victor Veitch, Alexander D'Amour, Steve Yadlowsky, and Jacob Eisenstein. 2021.
\newblock \href {https://openreview.net/forum?id=BdKxQp0iBi8} {Counterfactual
  invariance to spurious correlations in text classification}.
\newblock In \emph{Advances in Neural Information Processing Systems}.

\bibitem[{Wang et~al.(2019)Wang, Singh, Michael, Hill, Levy, and
  Bowman}]{wang2018glue}
Alex Wang, Amanpreet Singh, Julian Michael, Felix Hill, Omer Levy, and
  Samuel~R. Bowman. 2019.
\newblock \href {https://openreview.net/forum?id=rJ4km2R5t7} {{GLUE}: A
  multi-task benchmark and analysis platform for natural language
  understanding}.
\newblock In \emph{International Conference on Learning Representations}.

\bibitem[{Wang et~al.(2018)Wang, Dai, Kong, Ma, Erfani, Bailey, Xia, Song, and
  Zha}]{wang2018learning}
Y.~Wang, B.~Dai, L.~Kong, X.~Ma, S.~M. Erfani, J.~Bailey, S.~Xia, L.~Song, and
  H.~Zha. 2018.
\newblock Learning deep hidden nonlinear dynamics from aggregate data.
\newblock In \emph{Uncertainty in Artificial Intelligence (UAI)}.

\bibitem[{Wang and Culotta(2020)}]{Wang2020IdentifyingSC}
Zhao Wang and Aron Culotta. 2020.
\newblock \href {https://doi.org/10.18653/v1/2020.findings-emnlp.308}
  {Identifying spurious correlations for robust text classification}.
\newblock In \emph{Findings of the Association for Computational Linguistics:
  EMNLP 2020}, pages 3431--3440, Online. Association for Computational
  Linguistics.

\bibitem[{Wei and Zou(2019)}]{Wei2019EDAED}
Jason Wei and Kai Zou. 2019.
\newblock \href {https://doi.org/10.18653/v1/D19-1670} {{EDA}: Easy data
  augmentation techniques for boosting performance on text classification
  tasks}.
\newblock In \emph{Proceedings of the 2019 Conference on Empirical Methods in
  Natural Language Processing and the 9th International Joint Conference on
  Natural Language Processing (EMNLP-IJCNLP)}, pages 6382--6388, Hong Kong,
  China. Association for Computational Linguistics.

\bibitem[{Williams et~al.(2018)Williams, Nangia, and
  Bowman}]{williams2017broad}
Adina Williams, Nikita Nangia, and Samuel Bowman. 2018.
\newblock \href {https://doi.org/10.18653/v1/N18-1101} {A broad-coverage
  challenge corpus for sentence understanding through inference}.
\newblock In \emph{Proceedings of the 2018 Conference of the North {A}merican
  Chapter of the Association for Computational Linguistics: Human Language
  Technologies, Volume 1 (Long Papers)}, pages 1112--1122, New Orleans,
  Louisiana. Association for Computational Linguistics.

\bibitem[{Wolf et~al.(2020)Wolf, Debut, Sanh, Chaumond, Delangue, Moi, Cistac,
  Rault, Louf, Funtowicz, Davison, Shleifer, von Platen, Ma, Jernite, Plu, Xu,
  Le~Scao, Gugger, Drame, Lhoest, and Rush}]{Wolf2019HuggingFacesTS}
Thomas Wolf, Lysandre Debut, Victor Sanh, Julien Chaumond, Clement Delangue,
  Anthony Moi, Pierric Cistac, Tim Rault, Remi Louf, Morgan Funtowicz, Joe
  Davison, Sam Shleifer, Patrick von Platen, Clara Ma, Yacine Jernite, Julien
  Plu, Canwen Xu, Teven Le~Scao, Sylvain Gugger, Mariama Drame, Quentin Lhoest,
  and Alexander Rush. 2020.
\newblock \href {https://doi.org/10.18653/v1/2020.emnlp-demos.6} {Transformers:
  State-of-the-art natural language processing}.
\newblock In \emph{Proceedings of the 2020 Conference on Empirical Methods in
  Natural Language Processing: System Demonstrations}, pages 38--45, Online.
  Association for Computational Linguistics.

\bibitem[{Wu et~al.(2021)Wu, Ribeiro, Heer, and Weld}]{wu2021polyjuice}
Tongshuang Wu, Marco~Tulio Ribeiro, Jeffrey Heer, and Daniel Weld. 2021.
\newblock \href {https://doi.org/10.18653/v1/2021.acl-long.523} {Polyjuice:
  Generating counterfactuals for explaining, evaluating, and improving models}.
\newblock In \emph{Proceedings of the 59th Annual Meeting of the Association
  for Computational Linguistics and the 11th International Joint Conference on
  Natural Language Processing (Volume 1: Long Papers)}, Online. Association for
  Computational Linguistics.

\bibitem[{Zmigrod et~al.(2019)Zmigrod, Mielke, Wallach, and
  Cotterell}]{Zmigrod2019CounterfactualDA}
Ran Zmigrod, Sabrina~J. Mielke, Hanna Wallach, and Ryan Cotterell. 2019.
\newblock \href {https://doi.org/10.18653/v1/P19-1161} {Counterfactual data
  augmentation for mitigating gender stereotypes in languages with rich
  morphology}.
\newblock In \emph{Proceedings of the 57th Annual Meeting of the Association
  for Computational Linguistics}, pages 1651--1661, Florence, Italy.
  Association for Computational Linguistics.

\end{thebibliography}
\bibliographystyle{acl_natbib}

\newpage
\onecolumn
\appendix

\section{Toy Example Proof}
\label{app:proof}

In this section, we give the proof for Proposition~\ref{prop:main} for the toy example. For clarity, we also reproduce the statement of the proposition in this section:

\begin{prop}
\label{prop:duplicate}
Define the error for a model as $\ell(w) = \BE_{x\sim\sF}\pb{(w_{\text{inv}}^Tx - w^Tx)^2}$ where the distribution $\sF$ is the test distribution in which $x_r$ and $x_s$ are independent: $x_r \mid y \sim \sN(y\mu_r, \sigma_r^2I)$ and $x_s \sim \sN(0, I)$.

Assuming all variables have unit variance (i.e. $\sigma_r = 1$ and $\sigma_s$ = 1), $\|\mu_{r}\|$ = 1, and $\|\mu_{s}\| = 1$, we get $\ell(\hat{w}_{\text{inc}}) > \ell(\hat{w})$ if $\|\mu_{r1}\|^2 < \frac{1 + \sqrt{13}}{6} \approx 0.767$, where $\|\cdot\|$ denotes the Euclidean norm, and $\mu_{r1}$ is the mean of the perturbed robust feature $r_1$.
\end{prop}

\begin{proof}[Proof for Proposition~\ref{prop:duplicate}]
Given the definition of error we have,

\begin{align}
\label{eqn:error}
    \ell(\hat{w}) &= \BE_{x\sim\sF}\pb{(w_{\text{inv}}^Tx - \hat{w}^Tx)^2}
\end{align}

According to equation~\refeqn{inv}, we have ${w}_{\text{inv}} = \pb{\Sigma_r^{-1}\mu_r, 0}$ where

\begin{align}
    \Sigma_r = \text{Cov}(x_r,x_r) &= \BE_{x\sim\sD}\pb{x_rx_r^T} \nonumber\\
    &= \BE_{y\sim\sD}\pb{\BE_{x\sim\sD}\pb{x_rx_r^T|y}} \nonumber\\
    &= \BE_{y\sim\sD}\pb{I + y^2 \mu_r\mu_r^T} \nonumber\\
    &= I + \mu_r \mu_r^T
\end{align}

This gives us $\Sigma_r^{-1} = (I + \mu_r\mu_r^T)^{-1} = I - \alpha \mu_r\mu_r^T$ using the Sherman-Morrison formula since we have a rank-one perturbation of the identity matrix. Here $\alpha = \frac{1}{1 + |\mu_r|^2} = \frac{1}{2}$, giving $w_{\text{inv}} = \pb{\frac{\mu_r}{2}, 0}$.

Now note that according to equation~\refeqn{orig_model}, $\hat{w} = M^{-1}\mu$ where M, the covariance matrix can be written as a block matrix as in equation ~\refeqn{cov}. Hence we can formula for inverse of block matrix to get:

\begin{align}
    M^{-1} = \begin{bmatrix}
    I - \frac{1}{3}\mu_r\mu_r^T & -\frac{1}{3}\mu_r \mu_s^T\\
    -\frac{1}{3}\mu_s\mu_r^T & I - \frac{1}{3}\mu_s\mu_s^T
    \end{bmatrix}
\end{align}

Note that we have not shown the actual plugging in the formula of block matrix inverse, and simplifying but it is to verify that $M M^{-1} = I$. Therefore, we get

\begin{align}
    \hat{w} &= M^{-1}\mu \nonumber\\
    &= \begin{bmatrix}
    I - \frac{1}{3}\mu_r\mu_r^T & -\frac{1}{3}\mu_r \mu_s^T\\
    -\frac{1}{3}\mu_s\mu_r^T & I - \frac{1}{3}\mu_s\mu_s^T
    \end{bmatrix} \begin{bmatrix} \mu_r \\ \mu_s \end{bmatrix}\\
    &= \frac{1}{3}\mu
\end{align}

since $\|\mu_r\| = 1$ and $\|\mu_s\| = 1$. Plugging all these back into equation ~\refeqn{error}, we get:

\begin{align}
    \ell(\hat{w}) &= \BE_{x\sim\sF}\pb{(\frac{\mu_r^Tx_r}{2} - \frac{\mu^Tx}{3})^2} \nonumber\\
    &= \BE_{x\sim\sF}\pb{\frac{\mu_r^Tx_rx_r^T\mu_r}{4} + \frac{\mu^Txx^T\mu}{9} - \frac{\mu_r^Tx_rx^T\mu}{3}}
\end{align}

For the distribution $\sF$ we have, $\BE_{x\sim\sF}\pb{x_rx_r^T} = I + \mu_r\mu_r^T$ (since $x_r$ is distributed similarly in $\sD$ and $\sF$), $\BE_{x\sim\sF}\pb{x_rx^T} = \pb{I + \mu_r\mu_r^T, 0}$ (since $x_r$ and $x_s$ are independent in $\sF$) and $\BE_{x\sim\sF}\pb{xx^T} = \big(\begin{smallmatrix}I + \mu_r\mu_r^T & 0\\ 0 & I\\ \end{smallmatrix}\big)$. Plugging all these back and again using $\|\mu_r\| = 1$, $\|\mu_s\| = 1$, we get

\begin{align}
\label{eqn:orig_final}
    \ell(\hat{w}) &= \frac{1}{2} + \frac{2 + 1}{9} - \frac{2}{3} \nonumber\\
    &= \frac{1}{6}
\end{align}

For the incomplete edits, we have $\hat{w}_\text{inc} = [\Sigma_{r1}^{-1}\mu_{r1},0]$ where $\Sigma_{r1}^{-1} = (I + \mu_{r1}\mu_{r1}^T)^{-1} = I - \gamma\mu_{r1}\mu_{r1}^T$, $\gamma = \frac{1}{1 + \|\mu_{r1}\|^2}$ using the Sherman-Morrison formula again, since we have a rank-one perturbation of the identity matrix. This gives $\hat{w}_\text{inc} = \frac{1}{1 + \|\mu_{r1}\|^2} \pb{\mu_{r1}, 0}$. Note that $\BE_{x\sim\sF}\pb{x_rx_r^T} = I + \mu_r\mu_r^T$, $\BE_{x\sim\sF}\pb{x_{r1}x_{r1}^T} = I + \mu_{r1}\mu_{r1}^T$ and $\BE_{x\sim\sF}\pb{x_rx_{r1}^T} = \pb{I + \mu_{r1}\mu_{r1}^T, 0}^T$. Thus the error for incomplete edits is:

\begin{align}
\label{eqn:inc_final}
    \ell(\hat{w_{\text{inc}}}) &= \BE_{x\sim\sF} \pb{\frac{\mu_r^Tx_rx_r^T\mu_r}{4} + \frac{\mu_{r1}^Tx_{r1}x_{r1}^T\mu_{r1}}{(1 + \|\mu_{r1}\|^2)^2} - \frac{\mu_r^Tx_rx_{r1}^T\mu_{r1}}{1 + \|\mu_{r1}\|^2}} \nonumber\\
    &= \frac{1}{2} + \frac{\|\mu_{r1}\|^2}{1 + \|\mu_{r1}\|^2} - \|\mu_{r1}\|^2
\end{align}

Thus using equation ~\refeqn{orig_final} and ~\refeqn{inc_final}, we get $\ell(\hat{w}_{\text{inc}}) > \ell(\hat{w})$ if $3\|\mu_{r1}\|^4 - \|\mu_{r1}\|^2 -1 < 0$ which is exactly satisfied when $\|\mu_{r1}\|^2 < \frac{1 + \sqrt{13}}{6}$. 

\end{proof}

\section{Additional Experiments \& Results}

Here, we report more details on the experiments as well as present some additional results.

\subsection{Experiment Details}
\label{app:exp_details}
For NLI, models are trained for a maximum of 10 epochs, and for QA all models are trained for a maximum of 5 epochs (convergence is faster due to the larger dataset size). The best model is selected by performance on a held-out development set, that includes examples from the same perturbation type as in the training data. 

\subsection{Dataset Details}
\label{app:data_details}

The size of the training datasets and how they are constructed are described in Section~\ref{ssec:experiment}. Here, we give more details on the size of the various test sets used in the experiments. The size of the CAD datasets for the different perturbation types are given Table~\ref{tab:test_sizes} for both NLI and QA. Note that all test sets contain paired counterfactual examples, i.e. the seed examples and their perturbations belonging to that specific perturbation type. 

\begin{table}[t]
    \centering
    \begin{tabular}{ccc}
    \toprule
    Test Set & Size (NLI) & Size (QA)  \\
    \midrule
    \ctrltag{lexical}  & 406 & 314\\
    \ctrltag{resemantic} & 640 & 332\\
    \ctrltag{negation} & 80 & 268\\
    \ctrltag{quantifier} & 206 & 80\\
    \ctrltag{insert} & 376 & 118\\
    \ctrltag{delete} & 250 & - \\
    \bottomrule
    \end{tabular}
    \caption{Size of the tests sets corresponding to the different perturbation types for both NLI and QA. For QA, the number of examples in \ctrltag{delete} were extremely small and hence we do not use that perturbation type for QA.}
    \label{tab:test_sizes}
\end{table}

\subsection{Accounting for small dataset sizes}
\label{app:more_expt}

The experiments in Section~\ref{ssec:experiment} were run for 5 different random initializations, and we report the mean and standard deviation across the random seeds. For completeness, we also report results when using different subsamples of the SNLI dataset. Table~\ref{tab:nli_subsample} shows the mean and standard deviation across 5 different subsamples, along with the rest of the results which were presented in Section~\ref{sec:results_perturb_types}. We observe that even though there is variance in results across the different subsamples, majority of the trends reported in ~\ref{sec:results_perturb_types} are consistent across the different subsamples --- CAD performs well on aligned test sets, but does not necessarily generalize to unaligned test sets.

\begin{table*}[th]
\begin{small}
\centering
\begin{tabular}[t]{cccccccc}
\toprule
Train Data & All types & \ctrltag{lexical} & \ctrltag{insert} & \ctrltag{resemantic} &  \ctrltag{quantifier} & \ctrltag{negation} &   \ctrltag{delete}\\
\midrule
SNLI seed & 67.84\textsubscript{0.84} & 75.16\textsubscript{0.32} & 74.94\textsubscript{1.05} & \textbf{76.77}\textsubscript{0.74} & \textbf{74.36}\textsubscript{0.21} & \textbf{69.25}\textsubscript{2.09} &   65.76\textsubscript{2.34} \\
SNLI seed (subsamples) & 64.87\textsubscript{1.02} & 75.06\textsubscript{1.89} & 71.38\textsubscript{2.30} & 73.84\textsubscript{1.60} & 69.12\textsubscript{3.17} & 66.75\textsubscript{2.87} & 63.60\textsubscript{2.44}\\
\ctrltag{lexical} & 70.44\textsubscript{1.07} & \cellcolor{lightapricot}\textbf{81.81}\textsubscript{0.99} & 74.04\textsubscript{1.04}  & 74.93\textsubscript{1.16} & 72.42\textsubscript{1.58} &	68.75\textsubscript{2.16} &   67.04\textsubscript{3.00}\\
\ctrltag{insert} &  66.00\textsubscript{1.41} & 71.08\textsubscript{2.53} & \cellcolor{lightapricot}\textbf{78.98}\textsubscript{1.58}  & 71.74\textsubscript{1.53} & 68.15\textsubscript{0.88} & 57.75\textsubscript{4.54} &   68.80\textsubscript{2.71}\\
\ctrltag{resemantic} & \textbf{70.80}\textsubscript{1.68} & 77.23\textsubscript{2.35} & 76.59\textsubscript{1.12} & \cellcolor{lightapricot}75.40\textsubscript{1.44} & 70.77\textsubscript{1.04} & 67.25\textsubscript{2.05} &   \textbf{70.40}\textsubscript{1.54}\\
\bottomrule
\end{tabular}
    \caption{Results for the different perturbation types in NLI with multiple subsamples of the dataset. (\colorbox{lightapricot}{$\;$} denotes \emph{aligned test sets}). We observe that there is variance across different subsamples, but the majority of the trends reported in Section~\ref{sec:results_perturb_types} still hold true.}
\label{tab:nli_subsample}
\end{small}
\end{table*}

\begin{table*}[!th]
\begin{small}
\centering
\begin{tabular}[t]{cccccccc}
\toprule
Train Data & All types & \ctrltag{lexical} & \ctrltag{insert} & \ctrltag{resemantic} &  \ctrltag{quantifier} & \ctrltag{negation} &   \ctrltag{delete}\\
\midrule
SNLI seed & 71.41\textsubscript{0.40} & 79.90\textsubscript{1.00} & 78.08\textsubscript{0.49} & 79.84\textsubscript{1.17} & 75.92\textsubscript{1.17} & 77.25\textsubscript{2.42} &   70.88\textsubscript{0.68} \\
\ctrltag{lexical} & 73.10\textsubscript{0.56} & \cellcolor{lightapricot}\textbf{83.54}\textsubscript{0.91} & 77.28\textsubscript{0.64}  & 80.81\textsubscript{0.47} & 75.72\textsubscript{0.86} &	\textbf{78.00}\textsubscript{1.69} &   70.72\textsubscript{1.46}\\
\ctrltag{insert} &  72.91\textsubscript{0.54} & 80.39\textsubscript{0.88} & \cellcolor{lightapricot}\textbf{78.93}\textsubscript{0.66}  & 80.56\textsubscript{0.76} & \textbf{76.89}\textsubscript{0.84} & 77.25\textsubscript{2.66} &   71.43\textsubscript{2.40}\\
\ctrltag{resemantic} & \textbf{73.44}\textsubscript{0.33} & 81.23\textsubscript{0.64} & 77.97\textsubscript{0.51} & \cellcolor{lightapricot}\textbf{81.06}\textsubscript{0.49} & 76.60\textsubscript{1.42} & 75.75\textsubscript{2.03} &   \textbf{73.84}\textsubscript{1.25}\\

\bottomrule
\end{tabular}
    \caption{Results for the different perturbation types in NLI with larger dataset sizes, with 10\% of the data being the perturbations (\colorbox{lightapricot}{$\;$} denotes \emph{aligned test sets}).}
\label{tab:nli_more_seed}
\end{small}
\end{table*}

To account for the small dataset sizes, we also ran an experiment using the NLI CAD dataset analogous to the QA setup---using a larger number of SNLI examples (7000) and replace a small percentage of them (10\%) with perturbations of the corresponding perturbation type. We ensure that the original examples from which the perturbations were generated are also present in the dataset. Thus, all experiments will have much larger dataset sizes than before (7000 vs 1400), while still using counterfactual examples generated only by one specific perturbation type. The results for this experiment are reported in Table~\ref{tab:nli_more_seed}. We observe that CAD still performs best on aligned test sets but only marginally --- this happens since a large fraction of the dataset (90\%) is similar across all experiments. Although CAD performs worse on unaligned test sets than the aligned test sets, it does not necessarily perform worse than the SNLI baseline --- this happens since the larger number of seed examples will implicitly regularize the model from overfitting to that specific perturbation type.

\end{document}